%% file: main.tex
\documentclass[runningheads,a4paper]{llncs}

\input{head/00_settings}

\input{head/01_title}

\input{head/02_authors}

\begin{document}
    \mainmatter
    \maketitle
    \setcounter{footnote}{0} 


    \input{body/00_body}


    \bibliographystyle{vancouver}
    \bibliography{bibliography.bib}
\end{document}

%% file: head/00_settings.tex
\usepackage[top=2.5cm, bottom=2.5cm, left=2.5cm, right=2.5cm]{geometry}
\usepackage{amsmath}
\usepackage{graphicx}
\usepackage[x11names]{xcolor}
\usepackage[draft]{hyperref}  
\usepackage{caption}
\usepackage{subcaption}
\usepackage{url}
\usepackage{color}
\usepackage[superscript,biblabel]{cite}  
\usepackage{epsfig}
\usepackage{float}
\usepackage[framemethod=TikZ]{mdframed}  
\usepackage{dirtytalk}

\usepackage[nomargin,inline,marginclue,draft]{fixme}
\usepackage{cleveref}

\newcommand{\etal}{\textit{et al.\,}}
\newcommand{\myparagraph}{\vspace{1mm}\noindent}
\definecolor{panelcolor}{HTML}{F5DFD8}
\definecolor{panelheadingcolor}{HTML}{9D1536}
\newcommand{\std}[1]{\textcolor[gray]{0.3}{\pm#1}}

\newmdenv[
	leftmargin = 10pt,
	rightmargin = 10pt,
	backgroundcolor = panelcolor,
	linewidth = 0pt,
	skipabove = 10pt,
	skipbelow = 10pt
]{custombox}


%% file: head/01_title.tex
\title{(Predictable) Performance Bias in Unsupervised Anomaly Detection}
\titlerunning{(Predictable) Performance Bias in Unsupervised Anomaly Detection}


%% file: head/02_authors.tex
\author{
    Felix Meissen \inst{1,\thanks{\textit{Corresponding author}\\ \textit{Email address}: felix.meissen@tum.de\\ \textit{Phone}: +49\,176\,43440143 (Felix Meissen)}} \and
    Svenja Breuer \inst{2,3} \and
    Moritz Knolle \inst{1, 4} \and
    Alena Buyx \inst{2,5} \and
    Ruth M\"uller \inst{2,3} \and
    Georgios Kaissis \inst{1,6,8} \and
    Benedikt Wiestler \inst{7,9,10} \and
    Daniel Rückert \inst{1,8,10}
}
\authorrunning{F. Meissen, S. Breuer, M. Knolle \etal}

\institute{\scriptsize{
    Chair for AI in Healthcare and Medicine, Klinikum rechts der Isar der Technischen Universität München, Einsteinstr. 25, Munich, 81675, Germany \and
    Department of Science, Technology and Society, School of Social Sciences and Technology, and Technical University of Munich, Arcisstr. 21, Munich, 80333, Germany \and
    Department of Economics and Policy, School of Management, Technical University of Munich, Arcisstraße 21, 80333 Munich, Germany \and
    Konrad Zuse School of Excellence in Reliable AI, Munich Data Science Institute (MDSI), Walther-von-Dyck-Str. 10, Garching 85748, Germany \and
    Institute for History and Ethics of Medicine, School of Medicine, Technical University of Munich, Prinzregentenstraße 68, Munich, 81675, Germany \and
    Institute for Machine Learning in Biomedical Imaging, Helmholtz Munich, Ingolstädter Landstraße 1, 85764 Neuherberg, Germany \and
    Department of Diagnostic and Interventional Neuroradiology, Klinikum rechts der Isar, Ismaninger Str. 22, Munich, 81675, Germany \and
    Department of Computing, Imperial College London, London SW7 2AZ, U.K \and
    TranslaTUM, Center for Translational Cancer Research, Technical University of Munich, Ismaninger Str. 22, Munich, 81675, Germany \and
    BW and DR contributed equally as senior authors
}}

%% file: body/00_body.tex
\input{body/01_abstract}
\input{body/99_research_in_context}
\input{body/02_introduction}
\input{body/03_method}
\input{body/04_results}
\input{body/05_discussion}
\input{body/06_conclusion}
\input{body/07_additional_info}

%% file: body/01_abstract.tex
\par\noindent\rule{\textwidth}{0.4pt}
\renewcommand{\abstractname}{Summary}
\begin{abstract}

\mbox{}\\  

\textbf{Background}
With the ever-increasing amount of medical imaging data, the demand for algorithms to assist clinicians has amplified.
Unsupervised anomaly detection (UAD) models promise to aid in the crucial first step of disease detection.
While previous studies have thoroughly explored fairness in supervised models in healthcare, for UAD, this has so far been unexplored.

\vspace{2mm}\textbf{Methods}
In this study, we evaluated how dataset composition regarding subgroups manifests in disparate performance of UAD models along multiple protected variables on three large-scale publicly available chest X-ray datasets.
Our experiments were validated using two state-of-the-art UAD models for medical images.
Finally, we introduced a novel subgroup-AUROC (sAUROC) metric, which aids in quantifying fairness in machine learning.

\vspace{2mm}\textbf{Findings}
Our experiments revealed empirical \say{fairness laws} (similar to \say{scaling laws} for Transformers) for training-dataset composition: Linear relationships between anomaly detection performance within a subpopulation and its representation in the training data.
Our study further revealed performance disparities, even in the case of balanced training data, and compound effects that exacerbate the drop in performance for subjects associated with multiple adversely affected groups.

\vspace{2mm}\textbf{Interpretation}
Our study quantified the disparate performance of UAD models against certain demographic subgroups.
Importantly, we showed that this unfairness cannot be mitigated by balanced representation alone.
Instead, the representation of some subgroups seems harder to learn by UAD models than that of others.
The empirical \say{fairness laws} discovered in our study make disparate performance in UAD models easier to estimate and aid in determining the most desirable dataset composition.

\vspace{2mm}\textbf{Funding}
European Research Council Deep4MI

\end{abstract}
\par\noindent\rule{\textwidth}{0.4pt}

\subsection*{Keywords}
Artificial intelligence · Machine learning · Algorithmic bias · Subgroup disparities · Anomaly detection

%% file: body/99_research_in_context.tex
\begin{custombox}

\vspace{-6mm}
\subsection*{\textcolor{panelheadingcolor}{Research in context}}

\subsection*{Evidence before this study}

We searched PubMed, Scopus and Google Scholar for Machine Learning and Deep Learning studies related to \say{unsupervised anomaly detection}, \say{fairness}, \say{model bias}, and \say{intersectionality} before August 2023.
The list of publications was complemented by the authors' knowledge of the body of literature and suggestions from colleagues.
Several of these prior works have investigated the fairness of supervised classification models regarding protected attributes, such as gender, age, or race, and how this negatively affects patient care.
Dataset composition, fairness in population-wide studies, and the effect of intersectionality (considering multiple protected attributes) have been thoroughly investigated for this class of models.
These studies have found lower performances in under-represented or socio-economically disadvantaged patient groups, as well as disproportionate impacts in intersectional subgroups (i.e. individuals who share more than one sensitive trait).

\subsection*{Added value of this study}

Unsupervised anomaly detection (UAD) differs from supervised classification not only algorithmically but also regarding the context in which it is applied: the main application of UAD is pre-screening or triaging to support clinicians in handling increasing amounts of imaging data, and therefore has a distinct role.
Since UAD models learn the training data-generating distribution instead of an association between input and output labels, the effects for under-represented subpopulations are likely to be different from supervised models.
To this end, our work is the first to investigate and quantify fairness in UAD models.
Our experiments reveal empirical \say{fairness laws} -- linear relationships between dataset composition and subgroup performance that facilitate the \textit{a priori} estimation of subgroup performance.
The presented empirical results further show severe performance differences for subgroups even under balanced training data, suggesting that performance disparities in UAD can not be eliminated by equal data representation alone.
Lastly, we introduce a novel metric, subgroup-AUROC (sAUROC), to better quantify performance discrepancies for subgroups in machine learning models.

\subsection*{Implications of all the available evidence}

This study shows that performance bias is not limited to supervised classification models and further suggests additional care and rigour will be necessary in designing and deploying UAD algorithms to minimize excessive risk for misdiagnosis of different subpopulations.
This also implies that new UAD models generally need to be evaluated regarding their fairness.
We have further shown that performance bias behaves predictably in UAD.
The above-mentioned \say{fairness laws} render subgroup performance in UAD more predictable and can help to guide data collection towards more fair models.

\end{custombox}

%% file: body/02_introduction.tex
\section{Introduction} \label{sec:introduction}

In unsupervised anomaly detection (UAD), a machine learning (ML) model is trained to capture a distribution of training samples with the aim of being able to identify outliers/anomalies that do not stem from the distribution underlying the training data-generating process.
Within the medical domain, UAD serves as a valuable tool for detecting pathological samples while only requiring data derived from healthy patients without any disease-specific labels \cite{baur2021autoencoders,schlegl2019f,lagogiannis2023unsupervised}.
This approach allows for making use of the vast amounts of clinically unremarkable data acquired in hospitals on a daily basis.
In contrast to supervised methods, UAD, by construction, avoids the problem of class/label imbalance, making it well-suited to identify even rare anomalies, for which the collection of sufficient training data would otherwise present a challenge.
A UAD model $\Phi$ produces an anomaly score $a$ for a data point $x$.
Formally: $\Phi(x) = a$.
The model assigns higher anomaly scores to samples far from its learned distribution.


\myparagraph Recently, a lack of dataset diversity defined through, e.g. age, gender, or ethnicity, has been recognised as an important concern in medical imaging \cite{larrazabal2020gender} and beyond \cite{buolamwini2018gender}, as ML models trained on such data often provide biased predictions that result in poor performance on under-represented subgroups.
Furthermore, there is unequivocal evidence that this bias is harmful to patients: In a diverse dermatology dataset, Dansehjou \etal found that most state-of-the-art ML models used for skin cancer detection exhibited significantly lower performance than previously reported, particularly on dark skin tones.
In a clinical deployment setting, this can lead to delayed or missed diagnoses for patients with darker skin, potentially resulting in inadequate treatment and poorer outcomes \cite{daneshjou2022disparities}.
This issue becomes even more pronounced when taking intersectionality into account: Both Seyyed-Kalantari \etal and Stanley and colleagues independently reported that performance disproportionately worsens for patients belonging to multiple subgroups already adversely affected \cite{seyyed2021underdiagnosis,stanley2022disproportionate}.

\begin{figure}[ht]
    \centering
    \includegraphics[width=0.9\linewidth]{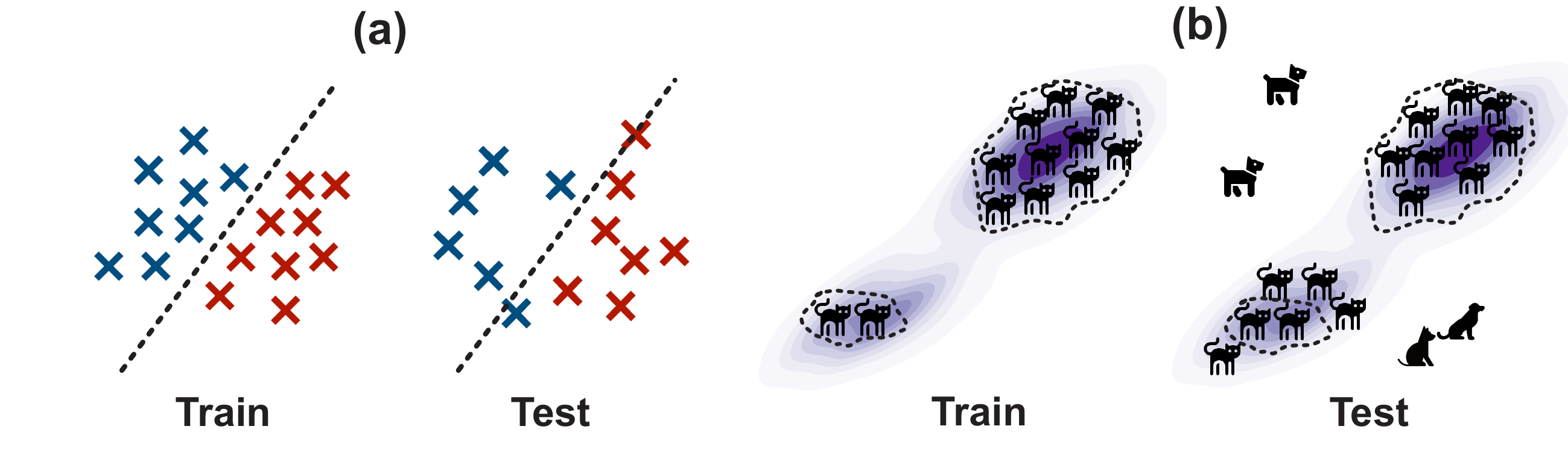}
    \caption{Supervised classification models (a) learn a mapping between input (samples) and (disease) labels and thus require annotated data. On the other hand, UAD models (b) learn to capture the distribution of the (healthy) training samples, here visualised as cats for exemplary purposes. If a subgroup is not adequately represented in the training data, the UAD model will assign data samples from that group higher anomaly scores, resulting in more false positives at test time.}
    \label{fig:anom_vs_supervised}
\end{figure}

\myparagraph For UAD models, which are trained to learn a \say{normal} distribution against which to contrast \say{anomalies} (see Fig. \ref{fig:anom_vs_supervised}), imbalanced training data is particularly challenging.
This is because subgroups observed less frequently in the training data lack sufficient representation for the UAD model to accurately learn their normal patterns, resulting in higher anomaly scores and, consequently, higher false-positive rates.
These can be detrimental in numerous ways, from unnecessary diagnostic tests and interventions to the potential for misdiagnosis or delayed diagnosis of true findings.
However, while the fairness of supervised ML models for clinical application has been increasingly studied \cite{larrazabal2020gender,seyyed2021underdiagnosis,petersen2022feature}, and the problem has been occasionally discussed in the UAD literature \cite{fairod,zhang2021towards}, a thorough empirical investigation of the matter for UAD models does not yet exist.

\myparagraph The main contribution of our study is a thorough investigation into the fairness of UAD models under different dataset compositions related to a protected attribute, including balanced scenarios. 
Notably, our work is the first to measure how the proportion of a subgroup in the training data of UAD models affects the resulting performance.
We operationalise our findings and introduce empirical \say{fairness laws} \footnote{Similar to \say{scaling laws} for transformers in natural language processing tasks \cite{kaplan2020scaling}} (i.e., a linear relationship between subgroup representation and performance), which help to identify the optimal dataset compositions for training fairer UAD models.
Finally, we introduced a new subgroup-AUROC (sAUROC) metric to better evaluate fairness in ML models.

%% file: body/03_method.tex
\section{Materials and Methods} \label{sec:methods}

\subsection*{Datasets}

We utilized three large public chest X-ray datasets to measure the fairness of UAD models regarding the protected attributes gender, age, and race: MIMIC-CXR-JPG (MIMIC-CXR) \cite{mimic-cxr}, ChestX-ray14 (CXR14) \cite{cxr14}, and CheXpert \cite{chexpert}.
MIMIC-CXR contains $371\,110$ chest X-rays from a cohort of $65\,079$ patients acquired at Beth Israel Deaconess Medical Center in Boston, Massachusetts, USA, between 2011 and 2016.
The CXR14 dataset, collected from the NIH Clinical Center in Bethesda, Maryland, USA, between 1992 and 2015, includes $112\,120$ frontal-view chest radiographs from $30\,805$ distinct patients.
The CheXpert database contains $224\,316$ chest radiographs of $65\,240$ patients acquired at Stanford Hospital between October 2002 and July 2017.

\myparagraph All three datasets contain structured diagnostic labels (12 labels for MIMIC-CXR and CheXpert, 13 for CXR14, excluding the \say{support devices} label) and a \say{no finding} or \say{normal} label marking the absence of any other identified diagnostic labels.
These labels were automatically derived from the associated radiology reports using natural language processing techniques.
In addition, demographic metadata about the patients' age (MIMIC-CXR: $60 \pm 18$, CXR14: $47 \pm 17$, CheXpert: $60 \pm 18$ years) and their gender (MIMIC-CXR: $47{\cdot}7\,\%$, CXR14: $43{\cdot}5\,\%$, CheXpert: $40{\cdot}6\,\%$ images of female and $59{\cdot}4\,\%$ of male patients \footnote{Categories like diverse/intersex/non-binary were missing in the data}) was available.
For MIMIC-CXR, additionally, the self-reported race was available from Johnson \etal \cite{johnson2020mimic}.
Here, $17{\cdot}3\,\%$ of the patients identified as \say{Black} and $62{\cdot}8\,\%$ as \say{White}.
While information about ethnicity was also available for CheXpert, the dataset sizes for the experiments described in Section \ref{sec:dataset_construction} were insufficient to adequately train anomaly detection models ($\sim 500$ images for training).

\subsection*{Dataset Construction and Inclusion Criteria} \label{sec:dataset_construction}

\begin{figure}[ht]
    \centering
    \includegraphics[width=\linewidth]{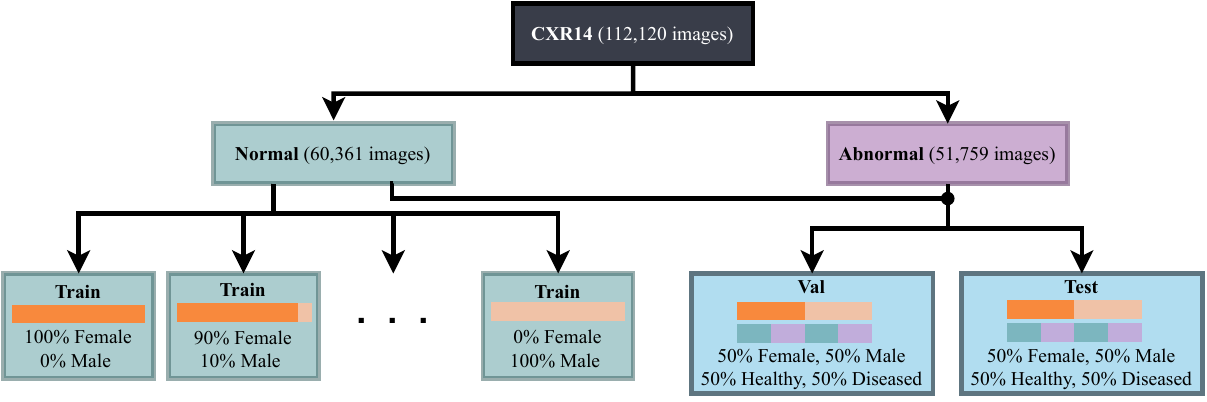}
    \caption{Exemplary flowchart of the procedure to generate training, validation, and test sets for CXR14.}
    \label{fig:dataset_construction}
\end{figure}

\myparagraph To ensure consistency in the inclusion criteria, distinct selection strategies were employed for all datasets.
We only considered frontal-view images without support devices and further excluded those with all labels marked as uncertain.
All images were center-cropped and resized to $128 \times 128$ pixels.

\myparagraph We constructed the training datasets using only the \say{no finding} and \say{normal} labels.
All other diagnostic labels were consolidated into a \say{diseased} label.
First, validation and test sets were randomly generated with equal representation of normal and abnormal classes and equal distribution of the protected attributes.
From the remaining normal data, we created training sets with varying proportions of the protected attribute subgroups (from $0\,\%$ to $100\,\%$).
In this process, the total number of training samples was held constant (e.g., the training set with $50\,\%$ \textit{male} and $50\,\%$ \textit{female} samples contains the same number of samples as the one with $10\,\%$ \textit{males} and $90\,\%$ \textit{females}).
The dataset construction for CXR14 is illustrated in Fig. \ref{fig:dataset_construction}.
For MIMIC-CXR and CheXpert, the procedure was analogous.
Importantly, there was no patient overlap between the training, validation, and test sets.

\myparagraph We selected two variables for the protected attribute race: \textit{white} and \textit{black}.
The group \textit{white} consists of the variable \say{WHITE}. We accumulated four variations of self-reported race into the group \textit{black}: \say{BLACK/AFRICAN AMERICAN}, \say{BLACK/CAPE VERDEAN}, \say{BLACK/AFRICAN}, and \say{BLACK/CARIBBEAN ISLAND}.
Further relevant subgroups, such as Asian or Latin, were excluded due to the small number of available images.
The categorization along the gender dimension was taken from the binary \textit{male} and \textit{female} values provided in the datasets.
For the separation between \textit{young} and \textit{old} patients, we opted for a data-driven approach, dividing the patient pool into three age groups based on the maximum age within the MIMIC-CXR dataset and removing the centre group to ensure a sufficient age gap between the patients in the two remaining groups.
This stratification resulted in individuals up to 31 years being classified as \textit{young}, while individuals aged 61 years and above were categorized as \textit{old} within our analysis.

\myparagraph We conducted additional experiments to examine fairness in intersectional groups, controlling for multiple protected attributes (gender, age, and race) in the MIMIC-CXR dataset.
We constructed random test sets for each possible combination of two protected attributes (i.e. \textit{male} and \textit{white}, \textit{male} and \textit{black}, \textit{female} and \textit{white}, etc.), each balanced in terms of both positive and negative samples and all considered protected attributes.
The remaining normal data was used to form the training set.
While this training set was unbalanced regarding the protected attributes, it approximates the population of the originating hospital.

\myparagraph Disease prevalence is unequal in the subgroups defined above.
This prevalence shift is known to cause disparities in many metrics, such as the receiver operating characteristics (ROC) curve \cite{glocker2023algorithmic}.
Since we intend to measure model performance bias in this study, we corrected for unequal prevalence while constructing our validation and test sets.
Since the prevalences in the datasets considered suffer from selection bias and, thus, likely do not represent the true prevalence in many real-world scenarios, we chose an arbitrary but consistent prevalence of $0.5$ in the test sets.

\subsection*{Statistics}

Since samples of under-represented subgroups are likely to yield higher anomaly scores and, consequently, are more often falsely flagged as positive (c.f. Section \ref{sec:introduction}), we considered predictive equality (or false positive error rate balance) as the most relevant measure to assess (group-)fairness in the context of UAD.

\myparagraph While the effect of generally higher or lower anomaly scores for a subgroup could potentially be partially mitigated \textit{post hoc} by selecting a unique threshold for each group, this solution can not be used in many cases where the association of a sample to a particular variation is unavailable, for example in retrospective cases.
The amount of required thresholds further grows combinatorially with the number of protected variables in intersectional subgroups, while the number of available samples to estimate this threshold shrinks simultaneously \cite{seyyed2021underdiagnosis}.
Thus, when evaluating an anomaly detection model's fairness regarding multiple subgroups, metrics such as the false positive rate at a minimally required true positive rate (FPR@x\%TPR) cannot be calculated separately for each group.
Instead, the threshold necessary to achieve the minimum TPR must be computed across the entire dataset, while the resulting FPRs should be determined individually for each subgroup \cite{glocker2023algorithmic}.
Similarly, the area under the receiver operating characteristics curve (AUROC) can not be computed for each group separately, as the minimum and maximum anomaly scores for both groups are likely significantly apart.
To alleviate this issue, we propose the subgroup-AUROC (sAUROC), which can be viewed as a threshold-free extension of the FPR@x\%TPR metric.

\myparagraph For a subgroup $s \in \mathcal{S}$ in the overall population $\mathcal{P}$, the sAUROC is calculated as follows: 
For every decision threshold $t$, the TPR is computed over the whole population, but the FPR only over the specific subgroup.
Mathematically, sAUROC can be described as:

\begin{equation}
    \mathrm{TPR}_\mathcal{P}(t) = \frac{\mathrm{TP}_\mathcal{P}(t)}{\mathrm{TP}_\mathcal{P}(t) + \mathrm{FN}_\mathcal{P}(t)} 
\end{equation}

\begin{equation}
    \mathrm{FPR}_s(t) = \frac{\mathrm{FP}_s(t)}{\mathrm{FP}_s(t) + \mathrm{TN}_s(t)} ,
\end{equation}

\myparagraph where $\mathrm{TP}(t), \mathrm{FN}(t), \mathrm{FP}(t), \mathrm{TN}(t)$ are the numbers of true positives, false negatives, false positives, true negatives at threshold $t$, respectively, and the subscripts $\mathcal{P}$ and $s$ denote if all samples are considered, or only the ones from the subgroup $s$, respectively.
Finally, the sAUROC is computed as

\begin{equation}
    \text{sAUROC}(s) = \int_0^1 \mathrm{TPR}_{\mathcal{P}} (\mathrm{FPR}_s^{-1}(x)) \mathrm{d}x .
\end{equation}

\myparagraph sAUROC paints a thorough picture of performance differences between subgroups of a population.

We ran each experiment ten times with different random seeds

\subsection*{Anomaly Detection Model}

In our experiments, we employed the Structural Feature Autoencoder (FAE) by Meissen \etal \cite{fae}, the best-performing method among contemporary state-of-the-art techniques in a recent comparative analysis by Lagogiannis \etal \cite{lagogiannis2023unsupervised}.
Model optimization was performed using the Adam optimizer, with a learning rate of $0{\cdot}0002$, for $10\,000$ iterations.
Each experiment was run with ten different random seeds to compute confidence intervals using a Gaussian-based approximation and to perform statistical significance tests.
To validate that our findings are not specific to the chosen model, we validated our results with the Reverse Distillation model (RD) by Deng \etal \cite{rd} -- the second-best model in Lagogiannis \etal \cite{lagogiannis2023unsupervised}.
The settings for both models were preserved at their default parameters.
These can be found in the Appendix, along with a short description of both models and the results for the RD model.

\subsection*{Ethics}

This research is exempt from ethical approval as the analysis is based on secondary data which is publicly available, and no permission is required to access the data.

\subsection*{Role of the funding source}

The funders had no role in study design, data collection, data analysis, data interpretation, or writing of the report. The authors had full access to all the data in the study and had final responsibility for the decision to submit for publication.

%% file: body/04_results.tex
\section{Results} \label{sec:results}

\begin{figure}[ht]
    \centering
    \includegraphics[width=\linewidth]{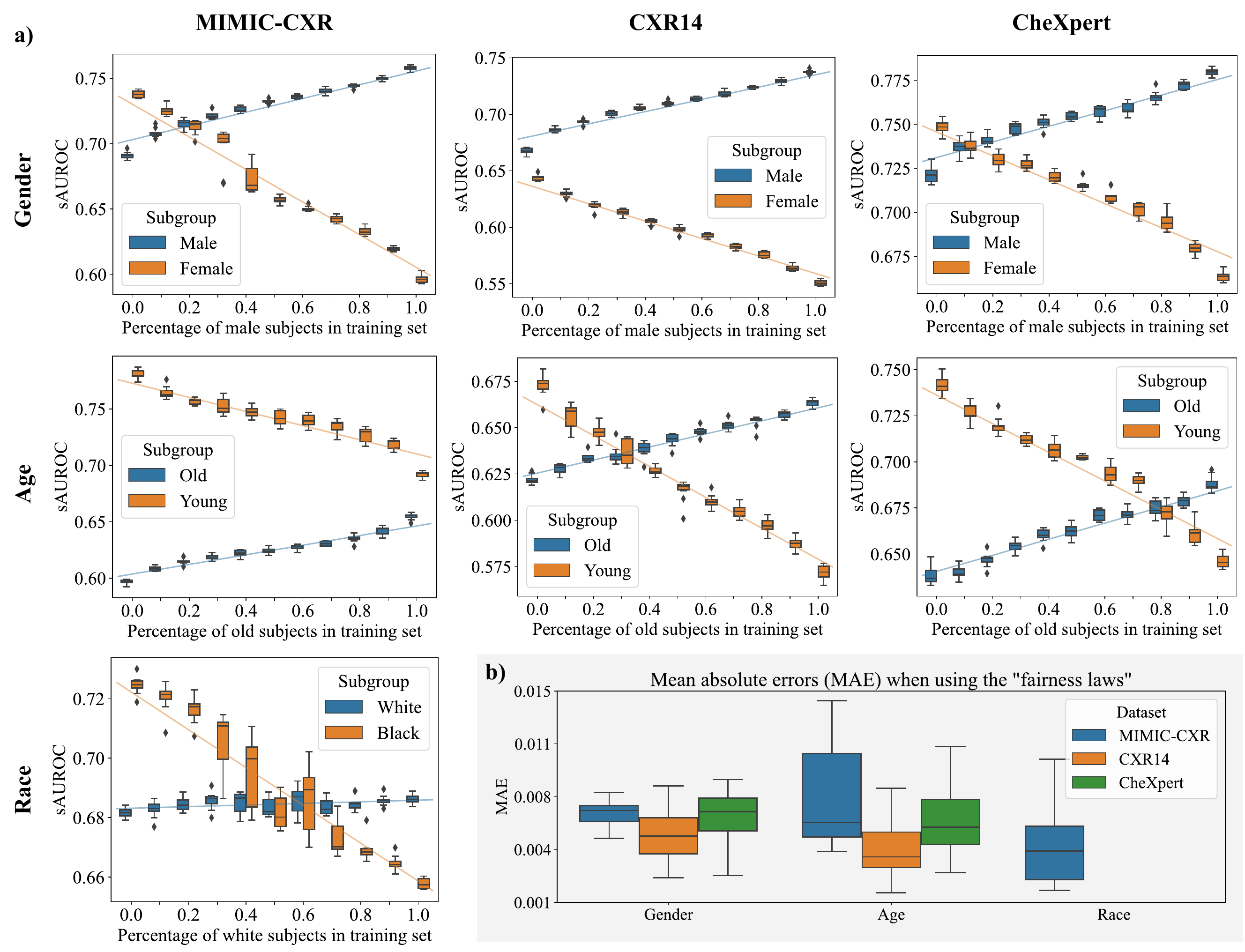}
    \caption{a) A linear relationship between the representation of a subgroup in the training dataset and its performance was observed across all datasets and subgroups. Equal representation of subgroups did not produce the most group-fair results. Experimental results for the FAE on the MIMIC-CXR, CXR14, and CheXpert datasets trained under different gender, age, or race imbalance ratios. Each box extends from the lower to upper quartile values of ten runs with different random seeds with a line at the median. Regression lines along the different imbalance ratios are additionally plotted. The exact numbers can be found in the Appendix.
    b) The mean absolute errors (MAE) between the real subgroup performances and those estimated using the ``fairness laws'' for each dataset and protected variable. Each box again shows the results over ten runs with different random seeds.}
    \label{fig:main_results}
\end{figure}


\subsection*{Subgroup representation and performance are linearly correlated}

\myparagraph Figure \ref{fig:main_results} shows a significant correlation between the representation of a subgroup in the training dataset and the subsequent performance for that subgroup (Pearson correlation coefficients of $0{\cdot}979\std{0{\cdot}011}$, $0{\cdot}971\std{0{\cdot}016}$, and $0{\cdot}682\std{0{\cdot}331}$ in the gender, age, and race experiments respectively, details in the Appendix).
This relationship was linear in all experiments, which allowed us to accurately estimate the sAUROC-performance for training with any composition of patients regarding a protected attribute, using linear interpolation from the extreme values ($0\,\%$ and $100\,\%$) -- with a mean absolute error of $0{\cdot}0061\std{0{\cdot}0029}$ for age, $0{\cdot}0063\std{0{\cdot}0015}$ for gender, and $0{\cdot}0045\std{0{\cdot}0025}$ for race.
Only in the race-controlled experiments on MIMIC-CXR the influence of the dataset composition on the performance outcome for patients from the \textit{white} group was much weaker (Pearson correlation coefficient $0{\cdot}398\std{0{\cdot}241}$).
The results of the RD model in the Appendix showed the same linear behaviour.

\subsection*{Unfairness exists with balanced training data}

\myparagraph The experiments in Figure \ref{fig:main_results} also revealed that \textit{male} subjects consistently received significantly higher scores across all datasets, even under balanced conditions, where both subgroups are equally represented in the training data (Welch's t-test, $N = 10$ different runs, $p < 0{\cdot}01$).
This pattern also held true for \textit{old} patients, except for the MIMIC-CXR dataset, where \textit{young} individuals obtained significantly higher scores ($p < 0{\cdot}01$).
Only when the protected variable was the patients' self-reported race, balanced training data did not lead to significant unequal performance in the FAE model ($p \ge 0{\cdot}01$).

\subsection*{Unfairness is amplified in intersectional subgroups}

\begin{figure}[ht]
    \centering
    \includegraphics[width=\linewidth]{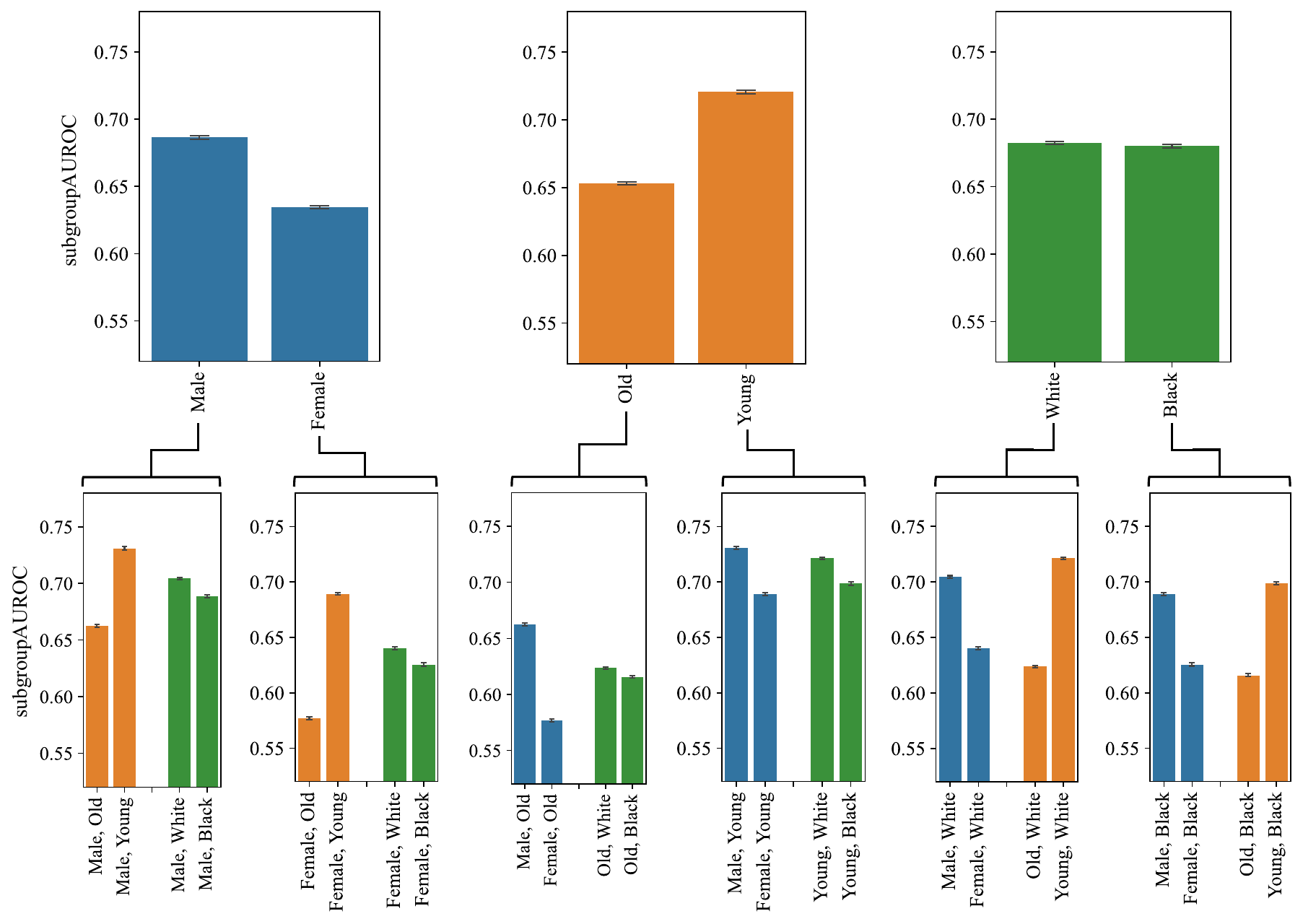}
    \caption{In the MIMIC-CXR dataset, representative of the Beth Israel Deaconess Medical Center, Boston, USA, diseases were detected better in \textit{male} than \textit{female} patients and in \textit{young} than \textit{old} patients. When considering a second demographic variable, these differences were amplified, e.g. the difference between \textit{male} and \textit{female} subjects is larger among older patients than younger ones. Top row: \textit{male} vs. \textit{female}, \textit{old} vs. \textit{young}, and \textit{white} vs. \textit{black}. Bottom row: intersectional subgroups. Each bar shows the mean and standard deviation over ten runs with different random seeds.}
    \label{fig:intersectional_results}
\end{figure}

\myparagraph Our intersectional experiments' outcomes are illustrated in Figure \ref{fig:intersectional_results}.
Given that the training dataset was not controlled for any protected variable, the findings presented here revealed potential unfairness within a population representative of the Beth Israel Deaconess Medical Center.
Here, \textit{male} patients received higher scores than \textit{female} patients, and \textit{young} patients achieved higher scores than their \textit{old} counterparts.
The performance of \textit{white} and \textit{black} patients appears to be equivalent.
This pattern is also consistently observable in the intersectional subgroups featured in the lower row of the Figure, although \textit{black} patients scored slightly lower in these cases.
Moreover, the score disparity ($\Delta$) between \textit{male} and \textit{female} patients was more pronounced among \textit{old} individuals compared to their \textit{young} counterparts ($\Delta 0{\cdot}085$ and $\Delta 0{\cdot}042$, respectively).
Similarly, the score disparity between \textit{old} and \textit{young} patients was larger among \textit{female} patients in comparison to \textit{male} patients ($\Delta 0{\cdot}112$ and $\Delta 0{\cdot}068$, respectively).

\subsection*{Naive AUROC fails to capture \say{fairness laws}}

\begin{figure}[ht]
    \centering
    \includegraphics[width=\linewidth]{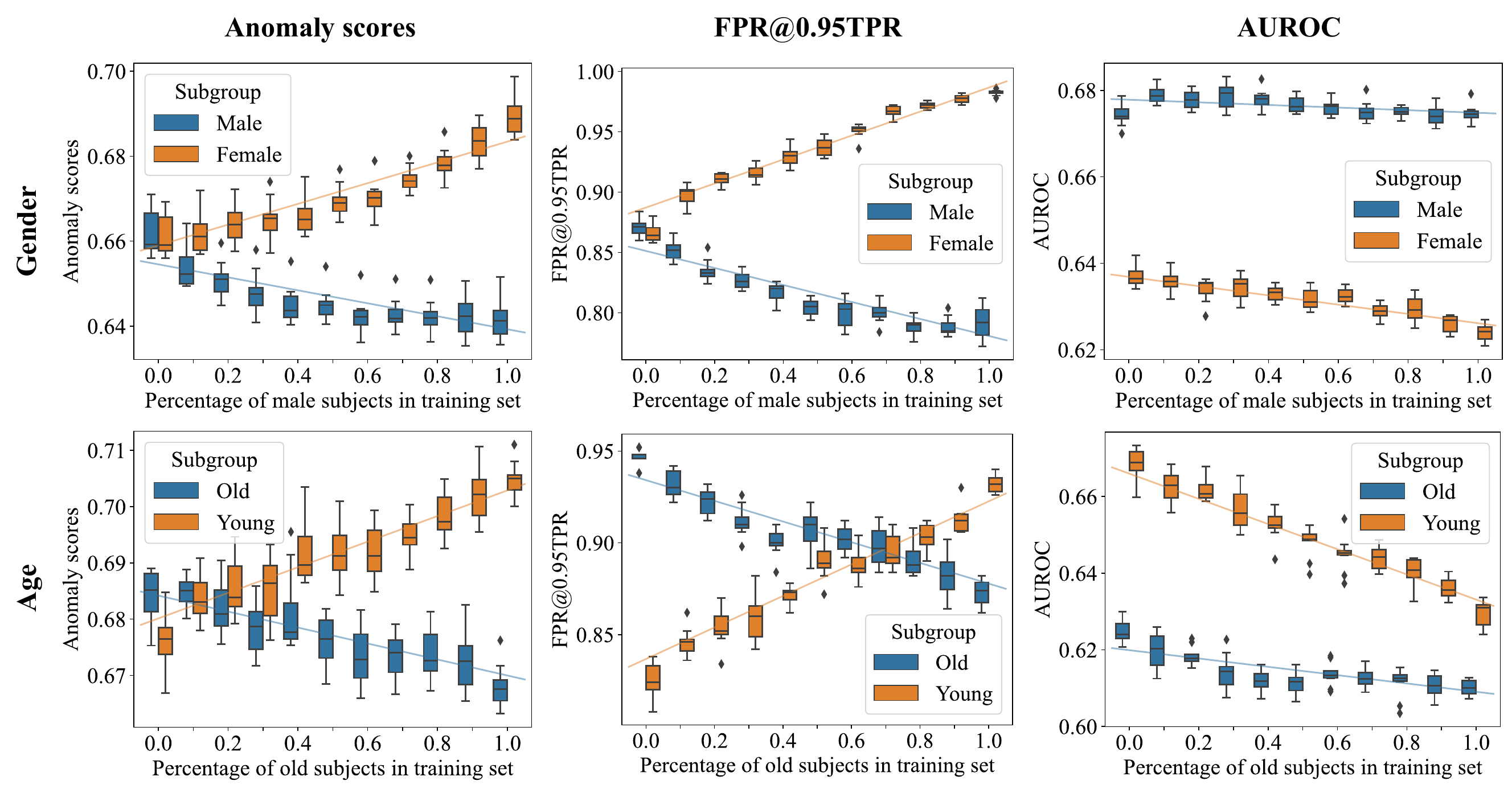}
    \caption{The representation of a subgroup in the training dataset had a strong influence on its anomaly scores, the false positive rate at a minimally required true positive rate, and our proposed sAUROC (c.f. Fig. \ref{fig:main_results}). Naive computation of AUROC did not capture this relationship. Anomaly scores (left), FPR@0·95TPR (middle), and naive AUROC (right) for different compositions of gender (top) and age (bottom) on the CXR14 dataset.}
    \label{fig:motivate_sAUROC}
\end{figure}

\myparagraph To put the sAUROC results in perspective, we additionally show the anomaly scores, FPR@\-0·95\-TPR, and naive AUROC for different dataset compositions on CXR14 in Figure \ref{fig:motivate_sAUROC}.
The anomaly scores for a subgroup increased as their representation in the training data shrank.
FPR@0·95TPR exhibited analogous behavior.
For AUROC, an increase in samples from one subpopulation did not improve scores for that group.
Instead, an increase in \textit{male} or \textit{old} patients resulted in similar or worse scores for all groups.

%% file: body/05_discussion.tex
\section{Discussion} \label{sec:discussion}

\myparagraph It has been shown that an embedded ethics and social science approach is helpful when analyzing complex, interdisciplinary problems like the one discussed in this work \cite{McLennan.2022,Breuer.2023}. 
We, therefore, drew on the interdisciplinary expertise of the authors in the discussion of the results, combining technical, social, and medical perspectives.


\myparagraph The experiments in Section \ref{sec:results} have unveiled \say{fairness laws} for UAD models: linear relationships between the representation of a subpopulation in the training data and the performance of that group (c.f. Figure \ref{fig:main_results}).
These relationships enable practitioners to accurately predict a subgroup's performance using only two points of measurement and linear inter- or extrapolation.
This implies that the optimal dataset combination under any fairness constraints can be easily estimated beforehand based on the above-described linear relationship.

\myparagraph However, like supervised methods \cite{stanley2022disproportionate}, UAD models suffered from compounding adverse effects in intersectional subgroups (c.f. Figure \ref{fig:intersectional_results}).
For example, the difference we found in model performance between \textit{male} and \textit{female} patients was larger in \textit{old} patients than in \textit{young} ones, resulting in \textit{old} \textit{female} patients being at a particular disadvantage.


\myparagraph Our experiments demonstrated substantial performance disparities among subgroups, even when they were equally represented in the training data.
For instance, \textit{male} patients on CXR14 consistently received significantly higher scores than their \textit{female} counterparts (\textit{male}: $0{\cdot}71$, \textit{female}: $0{\cdot}60$), and \textit{young} patients outperformed \textit{old} ones on MIMIC-CXR (\textit{young}: $0{\cdot}73$, \textit{old}: $0{\cdot}63$).
Notably, the dataset compositions that yield the most group-fair results -- as measured by predictive equality -- were often situated towards the extremes of the dataset composition spectra.
In CheXpert and MIMIC-CXR, optimal sAUROC parity was achieved with a $70\,\%$ and $80\,\%$ \textit{female} patient representation, respectively, whereas for CXR14, the composition that led to the most group-fair outcome did not include any \textit{male} samples.
This leads us to the hypothesis that there might be subgroup-specific differences that cause some subgroups to be easier to represent by the UAD model than others, as discussed by Nalisnick \etal \cite{nalisnick2018deep}.
Further potential reasons for this performance gap are summarized in a recent review by Petersen \etal \cite{petersen2023path}.
Among them are systematic labelling errors, as well as potential inherent task difficulty differences between groups.
Our findings, therefore, highlight the need for medical expertise in evaluating these models and their potential performance biases.


\myparagraph Disease detection is the central first step in the diagnostic process  \cite{kim2014fool}. 
Coupled with the increasing demand for medical imaging, UAD models fill a relevant clinical need.
In this pivotal role, unfairness in UAD, perhaps even more so than \say{downstream}, more specialized models, has significant potential to negatively affect patients.
Our experiments revealed that UAD models produce elevated false-positive rates for some subgroups (c.f. Figure \ref{fig:motivate_sAUROC}).
False positives can cause serious harm to patients, such as unnecessary follow-up tests (and costs) \cite{lafata2004economic}, harm from unnecessary treatment, and psychological distress \cite{brodersen2013long}, and can generally cause distrust in the model or ML techniques in general.


\myparagraph The results in Figure \ref{fig:motivate_sAUROC} show why sAUROC is a more suitable metric to measure the fairness of ML models.
The figure displays a linear relationship between anomaly scores and dataset composition, indicating that, as groups get less represented in the training data, they are flagged as \say{more anomalous} by the model.
These relationships were also reflected in the FPR@0·95TPR and our recently introduced sAUROC.
The naive, individual calculation of AUROC for each subpopulation, however, did not exhibit this expected pattern.
While the anomaly scores clearly showed disparities but were missing information about the resulting classification performance differences and FPR@0·95TPR is only a point measure considering only a single decision threshold, sAUROC painted a more comprehensive picture by considering all possible thresholds while capturing disparate performances.


\myparagraph The findings of our work need to be viewed with an awareness that the categories of human differences we are working with are complex and historically formed.
We could not find comprehensive information about how subjects were assigned labels of gender and race in the datasets.
This leaves us unclear about what social and biological aspects were included in these categories, important information for nuanced analyses that take into account how human bodies are shaped by both social and biological factors \cite{FaustoSterling2005sex,FaustoSterling2008race}.
Further, the labels of the datasets used for evaluation in this study are at risk of being biased.
Reasons for that reach from biases of medical professionals in the creation of the radiology reports \cite{fitzgerald2017implicit} to the automatic label extraction from these reports using a rule-based natural language processing system, which is known to generally contain high levels of label noise, especially in the oldest patient group \cite{zhang2022improving}.
Although UAD models only require minimal labels during training (healthy vs. diseased) and thus are presumably less susceptible to systematic labelling errors during training than supervised models, such errors might have an effect on our experimental results when they occur in the evaluation data.
Due to the small sample sizes of many ethnic subgroups in the available databases, our analysis of the race dimension was restricted to the categories \textit{white} and \textit{black} to guarantee meaningful insights of our results.
While we assume analogous effects on other racial categories like \textit{asian}, \textit{latin}, etc., the shortage of available data prevented us from empirically substantiating this hypothesis.
This limitation reflects the bias against under-represented racial groups that exists in current public medical data sets.



%% file: body/06_conclusion.tex

\myparagraph In conclusion, this study represents an effort to quantify fairness in UAD on a large scale, including the results of $1560$ trained models.
Our extensive experiments on various large-scale datasets and protected attributes confirmed that a demographic subpopulation's anomaly detection performance strongly depends on its representation in the training data and can be efficiently estimated, simplifying the task of identifying the fairest composition.
Our experiments further showed that disparate performance between two subgroups can not solely be explained by the under-representation of one subgroup.
Instead, some subgroups seemed to be harder to learn by the UAD models than others and, thus, were generally flagged as \say{more anomalous}.
While this study has found performance disparities existing along the three considered variables, likely, more of them exist (for example, people with disabilities).
Thus, enhancing model fairness is a significant yet unresolved requirement for the safe implementation of UAD models.
Towards this end, the sAUROC metric presented here is a relevant contribution, as it facilitates the quantification of performance bias in UAD models.
We emphasize that sAUROC is also relevant for supervised classification models where disparities between subgroups also manifest in TPR/FPR shifts \cite{glocker2023algorithmic,seyyed2021underdiagnosis}.
Considering the severe implications that over-diagnosis can have on both society and individual patients, we believe the quantification of existing bias mechanisms in this work presents a vital step towards a fairer future in healthcare.

%% file: body/07_additional_info.tex
\section*{Contributors}

F.M. conducted the experiments. F.M., S.B., M.K., and B.W. conceptualized the study, performed literature search, and validated and interpreted the experimental results. G.K., R.M., and D.R. acquired funding, provided resources, and supervised the study with regular feedback. All authors contributed to writing and critical revision of the manuscript.

\section*{Declaration of interests}

B.W. has received speaker honoraria from Novartis, Bayer and Philips. He holds a patent related to Apogenix APG101 for glioblastoma treatment and is a stockholder of the company "Need".

\section*{Acknowledgments}

This work was supported by the DAAD programme Konrad Zuse Schools of Excellence in Artificial Intelligence, sponsored by the Federal Ministry of Education and Research.
Daniel Rueckert has been supported by ERC grant Deep4MI (884622). Svenja Breuer, Ruth M\"uller and Alena Buyx have been supported via the project MedAIcine by the Center for Responsible AI Technologies of the Technical University of Munich, the University of Augsburg, and the Munich School of Philosophy.

\section*{Data sharing}

All data used in this work is publicly available.
The CXR14 dataset together with the patient demographic information can be downloaded from \url{https://nihcc.app.box.com/v/ChestXray-NIHCC/folder/36938765345}.
For the CheXpert dataset, analogous information is available from \url{https://stanfordmlgroup.github.io/competitions/chexpert/}.
The MIMIC-CXR dataset can be downloaded from \url{https://physionet.org/content/mimic-cxr-jpg/2.0.0/} with the corresponding patient demographic information available from \url{https://physionet.org/content/mimiciv/2.2/}.
All information to reproduce the exact results in this paper is available under an open source Apache 2.0 license in our dedicated GitHub repository \url{https://github.com/FeliMe/unsupervised_fairness}.